# Enhancing Retrieval for ESGLLM via ESG-CID: A Disclosure Content Index Finetuning Dataset for Mapping GRI and ESRS

**Shafiuddin Rehan Ahmed** **Ankit Parag Shah** **Quan Hung Tran** **Vivek Khetan**[†]
**Sukryool Kang** **Ankit Mehta** **Yujia Bao** **Wei Wei**
Center for Advanced AI, Accenture, Mountain View, CA, USA
{shafiuddin.r.ahmed,ankit.parag.shah,yujia.bao,wei.wei}@accenture.com
[†] Accenture Labs, San Fransisco, CA, USA
vivek.a.khetan@accenture.com

## Abstract

Environment, Social and Governance (ESG) reporting provides a diagnostic lens for evaluating a company's alignment with sustainability goals and stakeholder expectations, while also serving as an expression of its corporate identity and values. Frameworks like the Global Reporting Initiative (GRI) and the new European Sustainability Reporting Standards (ESRS) aim to standardize ESG reporting, yet generating comprehensive reports remains challenging due to the considerable length of ESG documents and variability in company reporting styles. To facilitate ESG report automation, Retrieval-Augmented Generation (RAG) systems can be employed, but their development is hindered by a lack of labeled data suitable for training retrieval models. In this paper, we leverage an underutilized source of weak supervision—the disclosure content index found in past ESG reports—to create a comprehensive dataset, `ESG-CID`, for both GRI and ESRS standards. By extracting mappings between specific disclosure requirements and corresponding report sections, and refining them using a Large Language Model as a judge, we generate a robust training and evaluation set. We benchmark popular embedding models on this dataset and show that fine-tuning BERT-based models can outperform commercial embeddings and leading public models, even under temporal data splits for cross-report style transfer from GRI to ESRS[1].

## 1 Introduction

ESG reporting serves as a diagnostic tool that enables structured self-assessment of a company's alignment with long-term sustainability goals and stakeholder expectations. It also is a comprehensive narrative that articulates the company's corporate identity, values, and, its impact to the world. The accelerating global climate crisis and increasing societal demands for corporate accountability have made ESG reporting a critical aspect of modern business. Natural Language Processing plays a pivotal role in understanding and drafting these long documents. Recent advancements in Large Language Models (LLMs) enable the analysis of vast amounts of textual data related to climate policies, sustainability reports, and environmental impact assessments (Vaghefi et al., 2023; Schimanski et al., 2024). By extracting insights from ESG reports, LLMs enhance transparency and inform stakeholders, driving data-driven decision-making in sustainability practices.

[1] huggingface.co/datasets/airefinery/esg_cid_retrieval

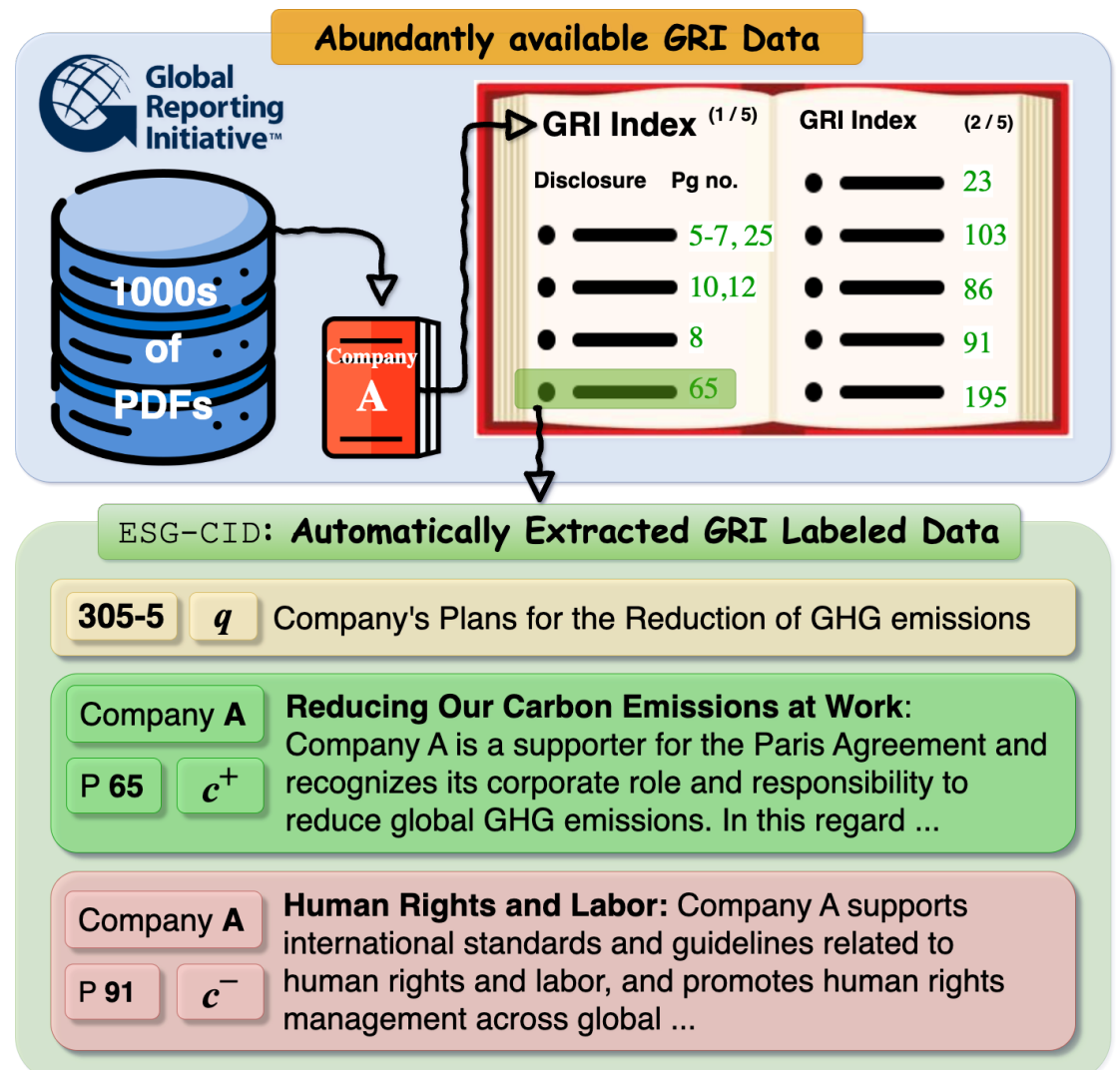

Figure 1: We extract content indices from GRI-compliant sustainability PDFs to create an ESG relevance dataset: ESG-CID. Each entry consists of a disclosure query ($q$), a relevant chunk ($c^+$) from the indexed page, and a randomly selected irrelevant chunk ($c^-$) from the rest of the document

Despite these advancements, generating comprehensive and standardized ESG reports remains a significant challenge. ESG documents are extensive—averaging 120 pages—and exhibit variability in reporting styles and structures among organizations. The lack of standardized and accessible ESG data can lead to greenwashing, obscures true risks, and impedes the effective allocation of resources toward sustainable investments and practices. Frameworks like the Global Reporting Initiative (GRI) and the new European Sustainability Reporting Standards (ESRS) aim to standardize ESG reporting, but automating this process requires effective Retrieval-Augmented Generation (RAG) systems. The development of such systems is hindered by a lack of labeled data suitable for training and evaluating retrieval models in the ESG domain.

The scarcity of labeled data arises mainly due to two factors: First, the considerable length of ESG reports makes manual annotation labor-intensive and time-consuming. Second, the lack of uniformity in reporting styles across different companies presents a challenge in creating datasets that generalize well. The combination of these factors makes it difficult to develop robust retrieval models needed for automating ESG reporting tasks.

In this paper, we leverage an underutilized yet readily available source of weak supervision: the **disclosure content index** found in past reports. We observed that GRI-compliant reports often include a content index linking specific disclosure requirements to corresponding sections or page numbers within the report. By extracting these mappings, we can generate large amounts of weakly supervised data that associates ESG disclosure queries with relevant text passages. To enhance the quality of this data, we use an LLM-as-a-judge to refine and validate the mappings. Additionally, it allows for an in-depth analysis of the standards' inter-relations providing insights on effectively using abundantly available past ESG data.

Using this dataset, we benchmark popular embedding models on the ESG retrieval task and explore the impact of fine-tuning. Our findings reveal that finetuning smaller BERT-based embedding models (`gte-large-en-v1.5`, `bge-large-en-v1.5`, `roberta-large`) can outperform commercial embedding models (`text-embedding-3-small`, `text-embedding-3-large`) and top-performing public models (`gte-Qwen2-1.5B-instruct`, `gte-Qwen2-7B-instruct`). Notably, our benchmark evaluates model performance under temporal data splits and cross-report style transfer from GRI to ESRS, demonstrating the generalizability of the fine-tuned models.

In summary, our contributions are as follows:

- We create the ESG-Content Index Dataset (`ESG-CID`), a dataset leveraging disclosure content indices from ESG reports to facilitate research in the ESG domain and support the development of retrieval models for standardized ESG reporting.
- We benchmark state-of-the-art embedding models on `ESG-CID`, highlighting their limitations in the ESG retrieval task out of the box and demonstrating the benefits of domain-specific fine-tuning.
- We conduct detailed analyses of model performance under temporal splits and cross-report style transfer, offering insights into the challenges and solutions for automating ESG report generation, particularly in the context of the new ESRS standards.

| Metric | Value |
| --- | --- |
| Unique Topics | 11 |
| Unique Sections | 112 |
| Total Datapoints | 1230 |
| Avg. Sections/Topic | 10 |
| Avg. Dataponts/Section | 11 |
| Sections with GRI Overlap | 99 |
| Sections without GRI Overlap | 13 |
| Sections GRI Overlap ratio | 0.88 |
| Datapoints with GRI Overlap | 648 |
| Datapoints without GRI Overlap | 582 |
| Datapoints GRI Overlap ratio | 0.53 |

Table 1: ESRS Statistics and Overlap with GRI. The table presents counts for unique topics, sections, and datapoints, along with their averages in the ESRS guidelines from the official GRI-ESRS interoperability data[2]. Section overlap is counted if at least one datapoint in the section overlaps with a GRI datapoint

## 2 Related Work

In our research, we build on the foundational work of GRI and the European Financial Reporting Advisory Group (EFRAG), which demonstrates the interconnection between the two standards—GRI and ESRS. Using their preliminary mapping, we illustrate the overlap between ESRS and GRI in Table 1. The table also presents statistics on unique topics, sections, and data points within ESRS, with significant overlaps highlighted in green. This overlap forms the basis of our approach, which is to leverage GRI data to meet ESRS standards.

The ESG domain has abundant public sustainability reports but lacks labeled data. Recent advancements in LLMs and PDF ingestion are bridg-ing this gap. Vaghefi et al. (2023) demonstrates the potential of LLMs to transform the ESG domain with a Climate-change query specific chat interface called *ChatClimate* powered by LLMs. More recent studies, such as *ChatReport* (Ni et al., 2023) and *ClimRetrieve* (Schimanski et al., 2024), focus on Question Answering within this domain through RAG. These studies, however, are limited by their focus on a narrow set of queries and evaluations based on only 10-20 documents. In contrast, our approach covers a broad spectrum of ESG framework requirements and queries, supported by extensive training and evaluation data.

[2] GRI-ESRS-Mapping.xlsx

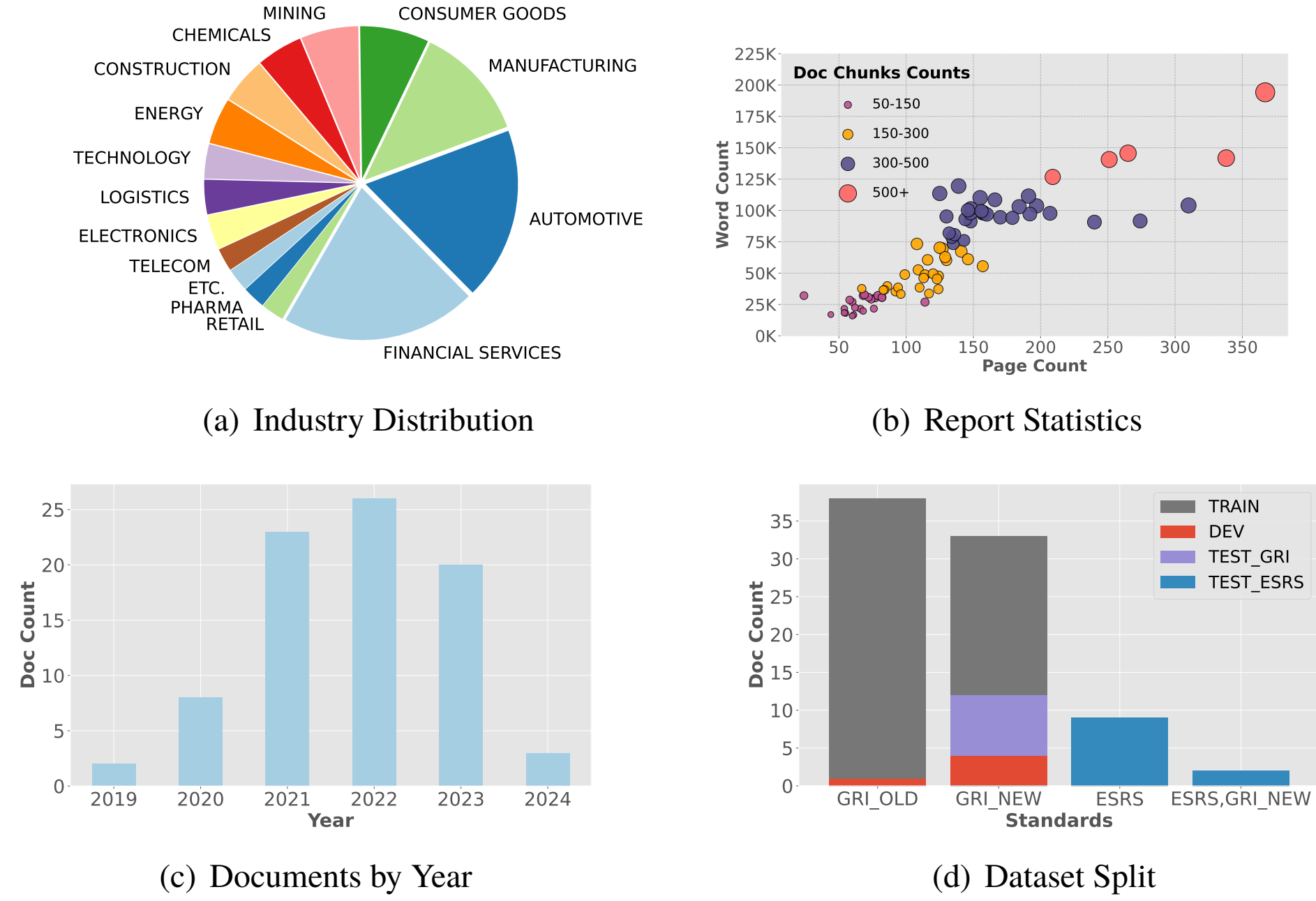

(a) Industry Distribution

(b) Report Statistics

(c) Documents by Year

(d) Dataset Split

Figure 2: Dataset characteristics and challenges: (a) Industry distribution, showcasing the diversity of reporting sectors. (b) Report statistics (page count vs. average word count per chunk, sized by chunk count), highlighting the variability in report length and chunk size, which pose challenges for retrieval models. (c) and (d): Dataset splits (Train, Dev, Test GRI, Test ESRS), illustrating the chronological approach and the out-of-domain ESRS test set.

Distant supervision is a key concept in low-resource model training (Quirk and Poon, 2017; Qin et al., 2018). Polignano et al. (2022) first proposed using the GRI content index as distant supervision for ESG annotations, focusing on table identification via Optical Character Recognition and its role in sentiment analysis. Our work extends this by linking ESRS and GRI frameworks and advancing representation learning through RAG-based automated content index creation.

RAG is a framework that enhances text generation by retrieving relevant external information, improving accuracy and contextual relevance in NLP tasks (Lewis et al., 2020; Jiang et al., 2023). However, most works on ESG domain rely on proprietary embeddings such as OpenAI, which are difficult to adapt to specific needs and pose privacy risks for company data. We enhance retrieval by fine-tuning on ESG-specific content indexes, exploring whether cost-efficient fine-tuning with high-quality data and smaller models can match more resource-intensive methods. We fine-tune various BERT-based models (both base and large) (Devlin et al., 2019; Liu et al., 2019; Li et al., 2023; Zhang et al., 2024; Xiao et al., 2023), leveraging the Model Test Evaluation Benchmark (MTEB; Muennighoff et al. (2022)) to identify the best-performing ones. Additionally, our study also evaluates ModernBERT (Warner et al., 2024) to further understand the impact of domain-specific fine-tuning on retrieval.

## 3 ESG-CID: Dataset Construction

In line with our goal to enhance ESG-specific retrieval systems, we first collected a comprehensive set of sustainability and annual reports from companies across various industries and regions. Utilizing a combination of automated web crawling and manual collection techniques, we gathered over 10,000 reports from 2018 to 2023. The automated collection leveraged databases such as the now-decommissioned GRI database and the SRN database (Donau et al., 2023). After filtering out duplicates and non-English reports, we retained ap-proximately 2,500 unique reports.

Out of these, around half adhered to the GRI standards, with a subset including the disclosure content index in a machine-readable format. We manually curated 73 GRI reports containing detailed content indices to form the primary dataset for our study. Additionally, we identified 11 reports from early adopters of the ESRS standards, which included ESRS content indices, enriching our dataset with cross-standard representations. The collected reports cover a diverse array of industries[3], predominantly from the financial, automotive, and manufacturing sectors (see Figure 2(a)).

### 3.1 Leveraging Content Indices for Weak Labeling

The disclosure content index serves as a structured bridge between the ESG standard requirements and the report content, providing an opportunity to create weakly labeled data without extensive manual annotation. Each content index lists the standard disclosure requirements (e.g., GRI or ESRS IDs and descriptions), along with references to the pages in the report where these disclosures are addressed.

As illustrated in Figure 2(b), the sustainability reports are significantly lengthy, averaging around 120 pages each, with the longest document exceeding 350 pages. Annotating such extensive documents is labor-intensive and impractical, especially when fine-grained annotations at the chunk or sentence level are considered. To address this challenge, we manually extracted only the content indices from the reports focusing only on these specific but crucial sections. Two experienced annotators, well-versed in ESG reporting and familiar with both GRI and ESRS standards, undertook this task. Their expertise ensured the accuracy and consistency of the extracted content indices.

Using the extracted content indices, we align the disclosure requirements with their corresponding page numbers in the reports. By automatically associating each standard query $q$ (i.e., the disclosure requirement) with the relevant sections of the report indicated by the page numbers, we generate a set of query-document pairs. The query is a standard disclosure requirement, and the document is the corresponding page content addressing that requirement. Leveraging this inherent structure allows us to create a weakly labeled dataset suitable for training and evaluating retrieval models.

[3]We provide the company name and year information of the reports of the dataset in §B

### 3.2 Creating Triplets for Embedding Models

To train and evaluate retrieval models in a contrastive learning framework, we construct triplets consisting of a query $q$, a positive (matched) chunk $c^+$, and a negative (unmatched) chunk $c^-$.

**Positive Chunks** We preprocess the PDF documents to segment them into manageable chunks (details in §D). The positive chunks $c^+$ are extracted from the pages referenced in the content index for each disclosure requirement. This ensures that $c^+$ contains information pertinent to the query $q$.

**Negative Chunks** For the negative samples $c^-$, we randomly sample chunks from the same report that are not associated with the given disclosure requirement. This assumes that these chunks are less relevant or irrelevant to the query, providing a contrastive signal for training.

### 3.3 Refining Labels with LLM Judgments

While the content indices provide page-level references, not all text within the referenced pages may directly address the disclosure requirement. To enhance the quality of our dataset, we employ Large Language Models (LLMs) as automated judges to assess the relevance of each chunk to the corresponding query.

We define a scoring function $s = \texttt{LLMScore}(q, c)$ that assigns a relevance score between 0 and 5 to each query-chunk pair. The LLM evaluates whether the chunk $c$ sufficiently addresses the disclosure requirement $q$. By applying a relevance threshold (e.g., $s \geq 3$), we filter out positive chunks that are not sufficiently relevant, thus improving the quality of the triplets.

This refinement step ensures that our dataset contains high-quality, relevant query-document pairs, enhancing the effectiveness of retrieval models trained or evaluated on this data[4].

### 3.4 Dataset Splitting for Real-World Evaluation

To simulate real-world scenarios, particularly the temporal evolution of ESG standards and the adoption of new reporting requirements, we strategically split our dataset based on report release years and reporting standards.

[4]Details on the LLM prompts and scoring criteria are provided in the §C

**Temporal Splitting** The 73 GRI reports are ordered chronologically. We allocate the 10 most recent reports released after 2020, which adhere to the updated `GRI-NEW` standards, to form the test set ($\texttt{TEST} - \texttt{GRI}$). The next 5 most recent reports are designated as the development set for hyperparameter tuning. The remaining 58 reports, primarily following the older `GRI-OLD` standards, constitute the training set as shown in Fig 2(d). This split emulates a scenario where models trained on earlier data are evaluated on newer standards, testing their ability to generalize over time.

**Cross-Standard Transfer** The 11 ESRS reports form a separate test set ($\texttt{TEST} - \texttt{ESRS}$), allowing us to assess the models' performance on a different but related standard. This setup facilitates the evaluation of cross-standard transferability and the models' adaptability to new reporting frameworks.

Organizing the dataset this way ensures our evaluations reflect the challenges faced in real-world applications, such as adapting to evolving standards and handling reports from different time periods.

# 4 Experimental Setup

## 4.1 Embedding Models

We benchmark the retrieval performance of several state-of-the-art embedding models, including both LLMs and lightweight BERT-based models (< 1B Params). The LLM-based embeddings comprise open-source models such as `gte-Qwen2-1.5B-instruct` and (Li et al., 2023), `gte-Qwen2-7B-instruct` (Li et al., 2023), which are known for their strong capabilities in capturing complex language representations. We also include commercial models from OpenAI, namely `text-embedding-3-small` and `text-embedding-3-large`.

In addition to the LLMs, we evaluate lightweight BERT-based models suitable for deployment in resource-constrained environments. These include `roberta-large` (Liu et al., 2019), `bge-large-en-v1.5` (Xiao et al., 2023), `ModernBERT-Large` (Warner et al., 2024) and `gte-large-en-v1.5` (Li et al., 2023; Zhang et al., 2024). We also compare their smaller `base` models thus offering balance between performance and computational efficiency. By comparing these models, we aim to understand the trade-offs between large-scale embeddings and more efficient alternatives in the ESG retrieval context.

## 4.2 Fine-tuning on `ESG-CID`

To enhance the domain-specific performance of the lightweight BERT-based models, we fine-tune them on the training split of our constructed dataset (`ESG-CID`). We utilize the standard Multiple Negatives Ranking Loss (Reimers and Gurevych, 2019) for contrastive learning using triplets consisting of a query, a positive chunk, and a negative chunk ($(q, c^+, c^-)$). Each query is associated with one relevant positive chunk and one irrelevant negative chunk, as detailed in Section 3.

The fine-tuning process spans five epochs and we pick the best checkpoint that achieves the lowest evaluation loss. Further training details are provided in the Appendix. The fine-tuned models using the entire training set are referred to by adding the suffix–`FT` to the model card (e.g., `roberta-large–FT`, `gte-large-en-v1.5–FT`, etc). Fine-tuned models trained by only using the `LLMScore`-curated training data have the suffix–$\texttt{FT}_{\texttt{LLM}}$. We hypothesize that fine-tuning will imbue these models with ESG-specific knowledge, improving their retrieval capabilities on domain-specific queries.

## 4.3 Evaluation Metrics

We evaluate the models using standard retrieval ranking metrics to assess their ability to retrieve relevant document chunks given a query. Since we do not directly label the relevant chunks for the disclosure and some chunks within the indexed page can be irrelevant, we slightly modify the evaluation. Given that the ground-truth is provided in the form of page numbers[5], we conduct the final ranking assessment based on relevant pages instead of chunks. This involves creating the assessment in a way that ranks page numbers using the metadata of the retrieved chunks.

The metrics calculated using the `ranx` library (Bassani, 2022) include:

**Recall@10**: Measures the proportion of relevant document pages retrieved in the top 10 chunks. We use '@10' to reflect the typical RAG use case that retrieves 10 documents.
**Mean Reciprocal Rank at 50 (MRR@50)**: Indicates how early the first relevant document page appears.
**Mean Average Precision at 50 (MAP@50)**: Averages precision scores at ranks where relevant document pages are found.

[5]assuming companies report their content index accurately and comprehensively

| Model | Size | TEST − GRI REC @10 | TEST − GRI MRR @50 | TEST − GRI MAP @50 | TEST − GRI NDCG @50 | TEST − ESRS REC @10 | TEST − ESRS MRR @50 | TEST − ESRS MAP @50 | TEST − ESRS NDCG @50 |
|---|---|---|---|---|---|---|---|---|---|
| gte-Qwen2-1.5B-instruct | 1.5B | 0.667 | 0.437 | 0.385 | 0.528 | 0.566 | 0.355 | 0.307 | 0.459 |
| gte-Qwen2-7B-instruct | 7B | 0.713 | 0.469 | 0.412 | 0.551 | 0.597 | 0.403 | 0.347 | 0.495 |
| text-embedding-3-small | | 0.684 | 0.459 | 0.405 | 0.545 | 0.546 | 0.336 | 0.284 | 0.439 |
| text-embedding-3-large | | 0.730 | 0.540 | 0.471 | 0.602 | 0.617 | 0.439 | 0.379 | 0.524 |
| *Frozen BERT-based Models* ❄️ | | | | | | | | | |
| roberta-base | 125M | 0.045 | 0.054 | 0.032 | 0.109 | 0.055 | 0.048 | 0.029 | 0.106 |
| BAAI/bge-base-en-v1.5 | 109M | 0.542 | 0.278 | 0.242 | 0.404 | 0.351 | 0.213 | 0.174 | 0.336 |
| Alibaba-NLP/gte-base-en-v1.5 | 137M | 0.603 | 0.366 | 0.313 | 0.465 | 0.461 | 0.277 | 0.225 | 0.390 |
| answerdotai/ModernBERT-Base | 150M | 0.112 | 0.078 | 0.056 | 0.165 | 0.157 | 0.103 | 0.072 | 0.194 |
| roberta-large | 355M | 0.146 | 0.107 | 0.08 | 0.203 | 0.161 | 0.110 | 0.077 | 0.189 |
| BAAI/bge-large-en-v1.5 | 335M | 0.608 | 0.373 | 0.325 | 0.475 | 0.435 | 0.257 | 0.212 | 0.374 |
| Alibaba-NLP/gte-large-en-v1.5 | 434M | 0.635 | 0.382 | 0.333 | 0.485 | 0.492 | 0.291 | 0.247 | 0.408 |
| answerdotai/ModernBERT-Large | 396M | 0.101 | 0.075 | 0.053 | 0.160 | 0.108 | 0.105 | 0.064 | 0.177 |
| *Fine-tuned BERT-based Models on entire data* (FT) | | | | | | | | | |
| roberta-base | | 0.77±.03 | 0.57±.02 | 0.51±.02 | 0.64±.02 | 0.59±.02 | 0.42±.02 | 0.35±.02 | 0.50±.02 |
| BAAI/bge-base-en-v1.5 | | 0.79±.01 | 0.61±.01 | 0.54±.01 | 0.66±.01 | 0.63±.01 | 0.45±.01 | 0.38±.00 | 0.53±.00 |
| Alibaba-NLP/gte-base-en-v1.5 | | 0.78±.01 | 0.60±.02 | 0.53±.02 | 0.65±.02 | 0.64±.03 | 0.45±.03 | 0.39±.02 | 0.53±.02 |
| answerdotai/ModernBERT-Base | –"– | 0.75±.01 | 0.54±.03 | 0.47±.02 | 0.61±.02 | 0.54±.02 | 0.37±.02 | 0.31±.02 | 0.46±.02 |
| roberta-large | | 0.78±.02 | 0.59±.03 | 0.52±.02 | 0.65±.02 | 0.60±.02 | 0.43±.02 | 0.36±.02 | 0.51±.02 |
| BAAI/bge-large-en-v1.5 | | 0.79±.02 | 0.59±.03 | 0.53±.03 | 0.65±.03 | 0.63±.03 | 0.46±.04 | 0.39±.04 | 0.54±.03 |
| Alibaba-NLP/gte-large-en-v1.5 | | 0.79±.01 | 0.59±.02 | 0.52±.02 | 0.65±.02 | 0.64±.02 | 0.45±.03 | 0.38±.03 | 0.53±.02 |
| answerdotai/ModernBERT-Large | | 0.78±.02 | 0.57±.02 | 0.50±.02 | 0.63±.02 | 0.57±.03 | 0.41±.03 | 0.34±.02 | 0.48±.02 |
| *Fine-tuned BERT-based Models on* LLMScore *filtered data* ($FT_{LLM}$) | | | | | | | | | |
| roberta-base | | 0.79±.01 | 0.59±.03 | 0.53±.03 | 0.65±.02 | 0.61±.03 | 0.43±.03 | 0.36±.03 | 0.51±.03 |
| BAAI/bge-base-en-v1.5 | | 0.79±.01 | 0.59±.02 | 0.53±.02 | 0.65±.02 | 0.63±.01 | 0.45±.02 | 0.39±.02 | 0.53±.01 |
| Alibaba-NLP/gte-base-en-v1.5 | | 0.79±.01 | **0.62**±.02 | 0.54±.02 | 0.66±.01 | 0.65±.02 | 0.46±.02 | 0.40±.02 | 0.54±.02 |
| answerdotai/ModernBERT-Base | –"– | 0.76±.04 | 0.56±.05 | 0.49±.05 | 0.62±.04 | 0.57±.06 | 0.39±.06 | 0.33±.05 | 0.48±.05 |
| roberta-large | | **0.80**±.01 | 0.61±.02 | 0.54±.03 | 0.66±.02 | 0.62±.03 | 0.45±.03 | 0.38±.03 | 0.53±.02 |
| BAAI/bge-large-en-v1.5 | | **0.80**±.01 | **0.62**±.02 | **0.55**±.01 | **0.67**±.01 | 0.65±.02 | 0.47±.03 | 0.40±.03 | 0.55±.02 |
| Alibaba-NLP/gte-large-en-v1.5 | | **0.80**±.01 | **0.62**±.02 | 0.55±.02 | **0.67**±.01 | **0.66**±.02 | **0.48**±.02 | **0.41**±.02 | **0.55**±.01 |
| answerdotai/ModernBERT-Large | | 0.79±.02 | 0.58±.04 | 0.52±.04 | 0.64±.03 | 0.59±.05 | 0.42±.05 | 0.35±.04 | 0.50±.04 |

Table 2: Overall effectiveness of the models on ESG-CID comparing the mean and std of the ranking metrics for the finetuned models on 5 different runs. The row corresponding to Alibaba-NLP/gte-large-en-v1.5 is highlighted as our best performing finetuned model, while OpenAI 's text-embedding-3-large serves as the best available baseline. Our best model outperforms the baseline by 7-8% on TEST − GRI and 3-4% on TEST − ESRS.

**Normalized Discounted Cumulative Gain at 50 (NDCG@50)**: Emphasizes the ranking positions of relevant document pages.

Performance is reported on both the GRI test split (TEST − GRI) and the ESRS test split (TEST − ESRS). It is noteworthy that the fine-tuned models were trained exclusively on the GRI training data and have not been exposed to any ESRS data, allowing us to evaluate their generalization capabilities across different ESG reporting standards.

### 4.4 Real-world Applicability: ESRS Content Indexing

Beyond standard retrieval metrics, we assess the practical utility of the models in constructing the ESRS content index within a company's report. According to ESRS, companies are required to provide structured disclosures in a tabular format. Our objective is to automate the extraction and indexing of relevant information from PDF reports according to each disclosure requirement.

In this task, given a document $D$ and a set of ESRS disclosure queries $Q = \{q_1, q_2, \ldots, q_n\}$, we aim to map each query $q_i$ to its corresponding page numbers in $D$. We experiment with reports from two companies—one in the automotive industry and one in agriculture—to capture diversity in reporting styles. We report the precision, recall and F1 of these mappings.

Each report $D$ is segmented into chunks, and for each disclosure query $q_i$, the model retrieves the *top-10* most relevant chunks from $D$. The retrieved chunks are then mapped back to their page numbers, using the LLMScore effectively constructing the content index. Evaluation is based on the accuracy of these mappings, reflecting the models' effectiveness in automating the ESRS content indexing process.

## 5 Results and Analysis

### 5.1 Benchmarking Pre-trained Embedding Models

Table 2 presents the retrieval performance of various state-of-the-art embedding models on the GRI

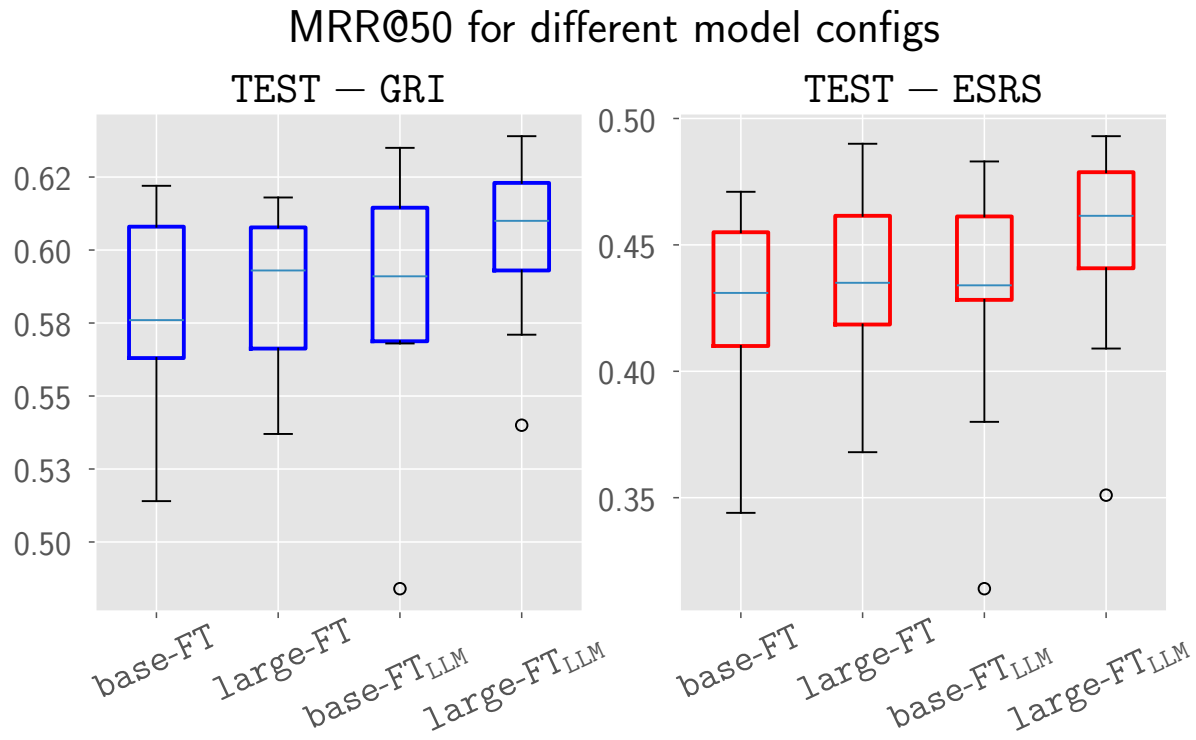

Figure 3: Box plot of the MRR@50 results from various fine-tuning runs (FT, $\text{FT}_{\text{LLM}}$) using base and large models. Each box represents the results from 20 different runs, comparing small and large BERT-based models in our experiments, with and without the use of LLMScore for filtering the training data.

and ESRS test sets. We show each finetuned model's aggregate performance on 5 different runs.

Firstly, we observe that most of the LLM-based embedding models demonstrate strong performance out of the box. For instance, the 1.5B parameter gte-Qwen2-1.5B-instruct embedding model achieves a Recall@10 of 0.667 without any domain-specific fine-tuning. Additionally, the open-source model gte-Qwen2-7B-instruct performs comparably to the commercial model text-embedding-3-large, highlighting the competitiveness of open-source solutions.

Secondly, LLM-based embedding models (listed in the first section of the table) significantly outperform the BERT-based embedding models (listed in the second section). This difference is attributed to the higher representational power and larger pretraining datasets of the LLM-based models, which enable better capture of semantic relationships in the ESG domain.

Thirdly, we note that the ESRS dataset presents a much greater challenge compared to GRI. There is a substantial performance degradation across models when evaluated on ESRS, indicating that ESRS retrieval tasks are more difficult.

## 5.2 Benchmarking Fine-tuned Embedding Models

We present the performance of our fine-tuned models in the last two sections of Table 2. While the original BERT-based models perform significantly worse than the LLM-based embeddings in their pretrained state, fine-tuning on our dataset results in substantial performance improvements. After fine-tuning, the BERT-based models not only close the gap but, in most cases, outperform the larger LLM-based embeddings.

Specifically, for the GRI test set, gte-large-en-v1.5–FT achieves improvements of over 5-6 percentage points across all ranking metrics. The other BERT-based models, both small and large, demonstrate consistent gains, outperforming the LLM-based models despite having fewer parameters. This showcases the effectiveness of fine-tuning on ESG-CID for enhancing model performance.

When evaluating the transfer performance to the ESRS test set, the fine-tuned models continue to perform significantly better than their pre-trained counterparts. Notably, the fine-tuned gte-large-en-v1.5–FT model outperforms the commercial baselines across all ranking metrics, despite not having been trained on any ESRS data. This suggests that fine-tuning on GRI data imparts transferable knowledge that generalizes to ESRS retrieval tasks to a great extent.

## 5.3 Impact of LLMScore Filtering

To understand the contribution of the LLMScore filtering step and see the difference in performance between the base and the large models, we plot the MRR@50 grouping the common runs. As shown in Figure 3, there is a consistent overall improvement when using the filtered data when compared to finetuning with entire data. This confirms that the LLM filtering helps to remove noise and im-prove the quality of the training data, leading to a more effective retrieval model. We also observe consistent (albeit small) improvements when using larger counterparts justifying their higher capacity for this GRI/ESRS retrieval task.

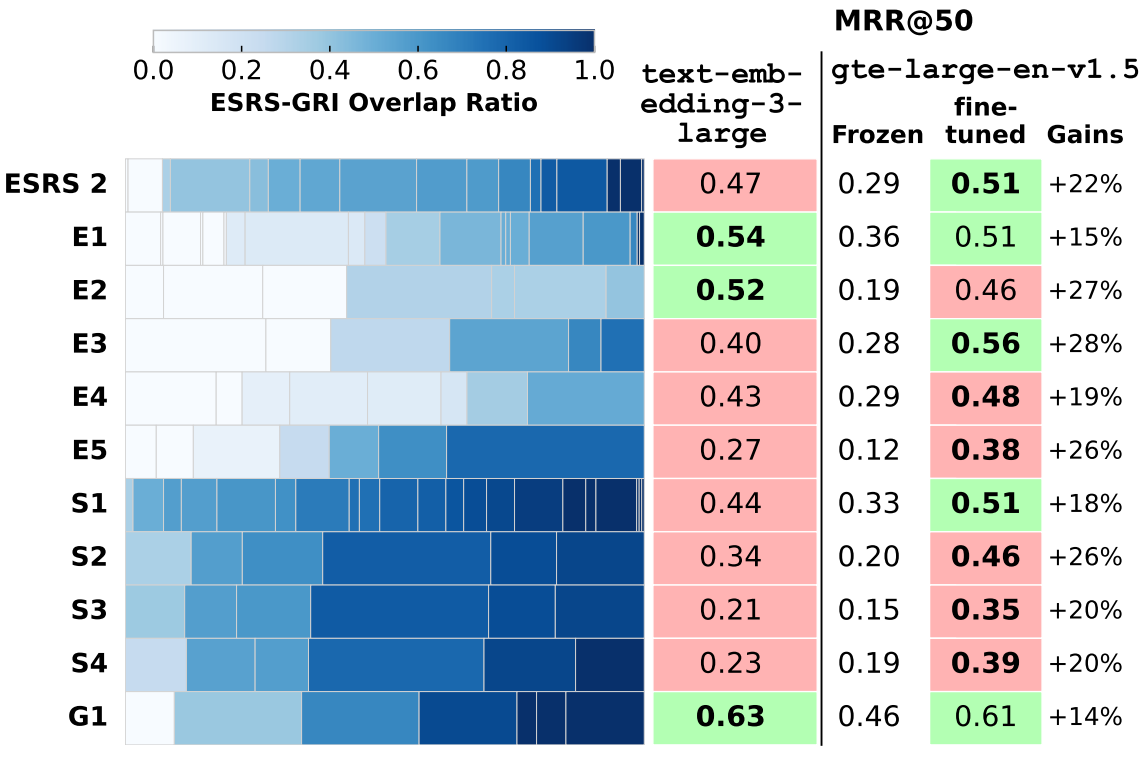

| MRR@50 | text-embedding-3-large | gte-large-en-v1.5 Frozen | gte-large-en-v1.5 fine-tuned | Gains |
|---|---|---|---|---|
| ESRS 2 | 0.47 | 0.29 | **0.51** | +22% |
| E1 | **0.54** | 0.36 | 0.51 | +15% |
| E2 | **0.52** | 0.19 | 0.46 | +27% |
| E3 | 0.40 | 0.28 | **0.56** | +28% |
| E4 | 0.43 | 0.29 | **0.48** | +19% |
| E5 | 0.27 | 0.12 | **0.38** | +26% |
| S1 | 0.44 | 0.33 | **0.51** | +18% |
| S2 | 0.34 | 0.20 | **0.46** | +26% |
| S3 | 0.21 | 0.15 | **0.35** | +20% |
| S4 | 0.23 | 0.19 | **0.39** | +20% |
| G1 | **0.63** | 0.46 | 0.61 | +14% |

Figure 4: ESRS-GRI overlapping datapoints grouped by topics (top to bottom). Sections within each topic are ordered by their overlapping ratio (left to right). The table on the right displays ranking scores, using the MRR@50 metric, comparing OpenAI embeddings, the frozen and the fine-tuned gte-large-en-v1.5 model. Scores from the better-performing model are boldened. Positive results (with MRR > 0.5) are highlighted in green, while negative results are highlighted in red.

### 5.4 Interplay between ESRS and GRI

To investigate the lower baseline scores observed in the ESRS test set, we conducted a detailed analysis of the overlap between ESRS topics and GRI standards. The heatmap in Figure 4 illustrates the overlapping sections, paired with the MRR@50 scores achieved by our best-performing model, `gte-large-en-v1.5`–FT$_{\text{LLM}}$, compared to the OpenAI baseline for each ESRS topic. We also include scores from the frozen counterpart to evaluate the performance gains from fine-tuning.

Our analysis reveals that the fine-tuned model consistently outperforms its frozen counterpart, with the most significant improvements observed in the `E2`, `E3`, `E5`, and `S2` topics, achieving gains of 26-27%. When compared to OpenAI's `text-embedding-3-large`, the fine-tuned model performs better in all but the `E1`, `E2`, and `G1` topics, with the maximum improvement of 16% observed in the `E3` topic, pushing the performance over the 50% MRR threshold.

However, certain topics, such as `E4` and `E5` (focusing on Biodiversity and Resource Use) remain challenging, as neither the large general-purpose model nor the fine-tuned model surpasses the 50% performance threshold. Similarly, topics from the Social category (`S2`, `S3`, and `S4`) show significant improvements from fine-tuning but still do not cross the threshold. In contrast, topics such as `ESRS 2` (General Disclosures), `E1`, `E3`, `S1`, and `G1` (Governance) demonstrate strong performance, indicating their suitability for automation. These topics exhibit high overlap with GRI, highlighting the potential to leverage existing GRI data to fine-tune retrieval systems for ESRS/CSRD-compliant reporting.

The problematic topics, highlighted in red, underscore areas where additional data collection and methodological refinement are necessary to improve mapping accuracy. Future work should focus on enhancing the GRI-ESRS correspondence or incorporating additional standards into the training set to further boost ESRS performance.

### 5.5 ESRS Content Indexing

Table 3 presents the results of ESRS content indexing, comparing the performance of our fine-tuned `gte-large-en-v1.5`–FT model with OpenAI embeddings. Our analysis reveals that `gte-large-en-v1.5`–FT$_{\text{LLM}}$ outperforms OpenAI embeddings in both the automotive and agricultural domains. Notably, our training set contains a substantial amount of automotive data but very few agricultural company reports, as illustrated in Figure 2(a). Despite this imbalance, `gte-large-en-v1.5`–FT$_{\text{LLM}}$ demonstrates emergent properties, generalizing well to the agricultural domain despite limited training data.

| Company | Model | Prec | Rec | F1 |
|---|---|---|---|---|
| Auto | text-embedding-3-large | 0.36 | 0.34 | 0.35 |
| | gte-large-en-v1.5 ❄ | 0.36 | 0.27 | 0.31 |
| | gte-large-en-v1.5–FT | 0.39 | 0.36 | 0.38 |
| | gte-large-en-v1.5–FT$_{\text{LLM}}$ | 0.39 | **0.40** | **0.40** |
| Agri | text-embedding-3-large | 0.62 | 0.42 | 0.50 |
| | gte-large-en-v1.5 ❄ | 0.67 | 0.40 | 0.50 |
| | gte-large-en-v1.5–FT | **0.69** | 0.43 | 0.53 |
| | gte-large-en-v1.5–FT$_{\text{LLM}}$ | 0.63 | **0.51** | **0.56** |

Table 3: Comparison of GTE and OpenAI models for content index generation on an Automotive (Auto) and an Agricultural (Agri) companies.

Interestingly, the inclusion of `LLMScore` reduces the precision of the RAG system. This suggests that models trained with LLM filtering may introduce hard relevant-looking false positives, thereby confusing the RAG system. Future work could address this issue through finer prompt tuning.

## 6 Conclusion

This paper addresses the critical need for scalable ESG information retrieval by leveraging disclosure content indices to align GRI and ESRS frameworks. By using content indices as a source of weak supervision, we developed a novel benchmark for ESG retrieval finetuning and showed our ESG models outperform strong baselines, such as OpenAI. Our results demonstrate GRI indices can effectively bootstrap models for ESRS compliance, achieving moderate transferability despite limited ESRS-specific data. The `LLMScore` filtering process further enhanced training data quality, enabling our models to generalize across evolving ESG standards. These findings highlight the practical benefits of structured indices in automating ESG reporting and compliance tasks. By harmonizing the GRI and ESRS frameworks, this research establishes a robust foundation for future inquiries into standard-agnostic capabilities, adaptability across regulatory frameworks, and holistic ESG reporting solutions.

## Limitations & Future Work

While our work lays a strong foundation for automated inter-framework ESG reporting and auditing, there are several limitations and areas for future research that we aim to address.

Firstly, the modest improvements between larger and smaller models suggest that our dataset may lack the size and diversity to fully exploit the capabilities of more complex models or the chosen samples for finetuning could be refined further being too noisy. Future research should focus on expanding and diversifying the dataset. This could include the incorporation of advanced techniques in automatic content index extraction from documents, leveraging recent advancements in PDF parsing and layout analysis on long documents (Saad-Falcon et al., 2023; Morio et al., 2024; Xie et al., 2025). Also, table reasoning through multi-agent refinement (Wang et al., 2024; Yu et al., 2025) could be explored to handle the diverse ESG reporting standards across different companies and frameworks more effectively. To address learning with noise, future work could investigate iterative training methodologies, such as multi-step training with hard negatives (Zhang et al., 2024) or using a cross-encoder as a re-ranker (Han et al., 2020) to filter out noise and harness a larger model's full potential.

Secondly, while retrieval is a crucial component of our RAG approach, it is not an endpoint. Future work should explore the automated generation of comprehensive sustainability reports from a wide array of a company's source documents. Current research (Ni et al., 2023; Wu et al., 2024), including ours, limits ESG analysis to a single document. Expanding this to include multiple documents such as financial reports, proxy statements, and annual reports would provide a more holistic and realistic approach to ESG reporting, reflecting the multifaceted nature of real-world data.

Lastly, our current work is restricted to the English language, which limits its applicability, especially given the diverse linguistic landscape of ESG reporting, particularly in Europe (Gutierrez-Bustamante and Espinosa-Leal, 2022). Future efforts should aim to extend this work to other languages, leveraging the availability of parallel corpora where companies report in multiple languages. This would not only enhance the accessibility and applicability of our models but also open up exciting avenues for analyzing the multilingual dependencies and nuances in ESG reporting.

## Ethics Statement

We highlight the ethical aspects related to the participation of annotators in research activities. We are committed to ensuring that our approach to data annotation is humane, respectful, and inclusive, as this not only enhances the quality of the datasets but also respects and preserves the dignity and rights of all participants.

## Acknowledgments

We thank the anonymous reviewers for their valuable comments which greatly helped improving our work. We would also like to extend our gratitude to Rozga Rhett, Dhruv Malik A, Kapil Mahajan and Jens Laue for their assistance throughout this project.

## Disclaimer

# A Hyperparameter settings

This section provides detailed information on the hyperparameter settings and training procedures used for fine-tuning the retrieval models (RoBERTa-large and GTE-large).

## A.1 Hyperparameter Optimization

We used a combination of prior work, best practices for transformer fine-tuning, and empirical evaluation on a small validation set (carved out from the training set) to select the hyperparameters. Specifically, we held out five documents from the training set to form a validation set. This validation set was used solely for checkpoint selection and is distinct from the development set used for model evaluation. The primary metric for checkpoint selection was ‘dev_cosine_accuracy‘, defined below.

## A.2 Training Arguments

Table 4 summarizes the key hyperparameters used for training. These settings were largely consistent across both RoBERTa-large and GTE-large, with the primary difference being the batch size due to GPU memory constraints.

We use saving and evaluation strategy based on the number of steps we take.

We used the ‘SentenceTransformerTrainingArguments‘ class from the ‘sentence-transformers‘ library to manage the training process. The key parameters are as follows:

- ‘output_dir‘: The directory where the trained models and checkpoints are saved. - ‘overwrite_output_dir‘: If ‘True‘, overwrites the contents of the output directory. - ‘num_train_epochs‘: The number of training epochs. We chose 5 epochs based on preliminary experiments, observing that performance plateaued after this point.

| Hyperparameter | RoBERTa-large | GTE-large |
|---|---|---|
| Training Epochs | 5 | 5 |
| Train Batch Size | 32 | 8 |
| Eval Batch Size | 32 | 8 |
| Warmup Ratio | 0.05 | 0.05 |
| FP16 | False | False |
| BF16 | False | False |
| Batch Sampler | No Duplicates | No Duplicates |
| Eval Steps | 50 | 50 |
| Save Steps | 50 | 50 |
| Save Total Limit | 5 | 5 |
| Logging Steps | 20 | 20 |
| Learning Rate | 5e-5 | 5e-5 |
| Load Best Model | True | True |
| Weight Decay | 0.01 | 0.01 |
| Metric for Best Model | ‘cosine accuracy‘ | ‘cosine accuracy‘ |
| DDP Find Unused Params | False | False |

Table 4: Hyperparameter settings for fine-tuning RoBERTa-large and GTE-large.

- ‘per_device_train_batch_size‘: The batch size per GPU during training. We used a batch size of 32 for RoBERTa-large and 8 for GTE-large due to GPU memory limitations. - ‘per_device_eval_batch_size‘: The batch size per GPU during evaluation. - ‘warmup_ratio‘: The proportion of training steps used for a linear warmup of the learning rate. - ‘fp16‘ and ‘bf16‘: These were set to false due to hardware constraints. - ‘batch_sampler‘: We used the ‘NO_DUPLICATES‘ batch sampler, which ensures no duplicate examples within a batch. - ‘eval_strategy‘ and ‘eval_steps‘: Evaluation was performed every 50 training steps. - ‘save_strategy‘ and ‘save_steps‘: Model checkpoints were saved every 50 training steps. - ‘save_total_limit‘: Limited to 5 checkpoints to conserve disk space. - ‘logging_steps‘: Training statistics were logged every 20 steps. - ‘learning_rate‘: The initial learning rate for the AdamW optimizer was set to 5e-5. - ‘load_best_model_at_end‘: If ‘True‘, loads the model checkpoint with the best performance on the validation set at the end of training. - ‘weight_decay‘: The weight decay parameter for the AdamW optimizer. - ‘metric_for_best_model‘: The metric used for best model checkpoint selection was ‘eval_gri-chunk-dev_cosine_accuracy‘. - ‘ddp_find_unused_parameters‘: Set to ‘False‘ since distributed data parallel (DDP) training was not used.

## A.3 Loss Function and Evaluation

The loss function used was ‘MultipleNegativesRankingLoss‘ from the ‘sentence-transformers‘ library. This loss function is designed for contrastive learning, ensuring that similar pairs (query and positive chunk) have higher similarity scores than dissimilar pairs (query and negative chunk). Each batch considered all other examples as negatives.

For development set evaluation, we used the ‘TripletEvaluator‘ from ‘sentence-transformers‘. The ‘TripletEvaluator‘ takes three lists as input:

- ‘anchors‘: A list of query examples. - ‘positives‘: A list of relevant chunks. - ‘negatives‘: A list of irrelevant chunks.

The evaluator computes the cosine similarity between anchor-positive and anchor-negative embeddings and calculates the ‘cosine_accuracy‘ metric.

### A.4 Cosine Accuracy Metric

The ‘eval_gri-chunk-dev_cosine_accuracy‘ metric is calculated as follows:

1. Compute the cosine similarity between the query embedding and the positive chunk embedding: ‘sim_pos = cosine_similarity(M(q), M(c+))‘. 2. Compute the cosine similarity between the query embedding and the negative chunk embedding: ‘sim_neg = cosine_similarity(M(q), M(c-))‘. 3. Count the number of triplets where ‘sim_pos > sim_neg‘. 4. Compute ‘cosine_accuracy‘ as the percentage of triplets where the positive chunk has a higher cosine similarity to the query than the negative chunk.

This metric reflects the model’s ability to rank relevant chunks higher than irrelevant chunks.

### A.5 Training Procedure

The models were trained using ‘MultipleNegativesRankingLoss‘, which is well-suited for contrastive training. Triplets of (query, positive chunk, negative chunk) were constructed, ensuring each query had one associated positive and one negative chunk. No significant overfitting was observed during the five training epochs.

## B Company Information

See Table 5 for the company names and publication years of the ESG reports used in `ESG-CID`.

## C `LLMScore`Prompt Details

Below is the prompt used for ‘LLMScore‘, which leverages a Large Language Model (LLM) to assess the relevance of a text chunk to a given query, both extracted from an ESG report. The LLM is instructed to provide a numerical score on a scale of 0 to 5, reflecting the degree of relevance. See Figure 5 for further details.

**`LLMScore` Prompt**

**Given the following `[query]`, and a `[text chunk]` from an ESG report, please rate the relevancy of the chunk to the disclosure on a scale of 0-5, in terms of being able to provide evidence for the disclosure. Provide higher rating if the chunk has enough evidence to answer the query.**

- The output should be a single number between 0 and 5. 0 means not relevant at all, 5 means highly relevant.
- The output should be an integer

```
[query]
{disclosure}
[text chunk]
{chunk}
```

**Relevancy Score (1-5):** `<YOUR ANSWER HERE>`

Figure 5: Prompt for `LLMScore`

## D PDF Preprocessing

For the ingestion of long sustainability PDF documents, we adopt the popular `PyMUPdfLoader` library with scalability in mind. After extracting the text from each page of the report we perform the following steps:

1. **Newline Removal:** Remove newline characters to produce continuous text.
2. **Chunking:** Partition the text on a pagewise basis into segments of 2048 characters.
3. **Overlap:** Apply an overlap of 512 characters between contiguous chunks to preserve context.

Formally, for a given PDF document $d \in \mathcal{D}$, the loader produces a set of text chunks:

$$\mathcal{C}(d) = \{c_1, c_2, \dots, c_n\},$$

where each chunk $c_i$ is a sequence of 2048 characters (with a 512-character overlap with $c_i$ and $c_{i+1}$). These chunks serve as the basic units for further processing in our pipeline.

| DOCUMENT NAME | COMPANY | YEAR | INDUSTRY_CLUSTER | STANDARDS | SPLIT |
|---|---|---|---|---|---|
| FORD_2024 | FORD | 2024 | AUTOMOTIVE | ESRS | TEST_ESRS |
| HYUNDAI_2019 | HYUNDAI | 2019 | AUTOMOTIVE | GRI_OLD | TRAIN |
| HYUNDAI_2020 | HYUNDAI | 2020 | AUTOMOTIVE | GRI_OLD | TRAIN |
| HYUNDAI_2021 | HYUNDAI | 2021 | AUTOMOTIVE | GRI_OLD | TRAIN |
| HYUNDAI_2022 | HYUNDAI | 2022 | AUTOMOTIVE | GRI_OLD | DEV |
| HYUNDAI_2022_A | HYUNDAI | 2022 | AUTOMOTIVE | GRI_OLD | TRAIN |
| HYUNDAI_2023 | HYUNDAI | 2023 | AUTOMOTIVE | ESRS, GRI_NEW | TEST_ESRS |
| HYUNDAI_2024 | HYUNDAI | 2024 | AUTOMOTIVE | ESRS, GRI_NEW | TEST_ESRS |
| KIA_2024 | KIA | 2024 | AUTOMOTIVE | GRI_NEW | TEST_GRI |
| SKODA_2023 | SKODA AUTO | 2023 | AUTOMOTIVE | ESRS | TEST_ESRS |
| TOYOTA_2023 | TOYOTA | 2023 | AUTOMOTIVE | GRI_NEW | TEST_GRI |
| TRAIN_18 | Nissan Motor Corporation | 2022 | AUTOMOTIVE | GRI_OLD | TRAIN |
| TRAIN_186 | Nissan Motor Corporation | 2021 | AUTOMOTIVE | GRI_OLD | TRAIN |
| TRAIN_25 | Geely Automobile Holdings | 2022 | AUTOMOTIVE | GRI_NEW | TRAIN |
| TRAIN_22 | Benteler Group | 2022 | AUTOMOTIVE | GRI_NEW | TRAIN |
| TRAIN_123 | SKC | 2023 | CHEMICALS | GRI_NEW | TRAIN |
| TRAIN_294 | NOVA Chemicals | 2021 | CHEMICALS | GRI_OLD | TRAIN |
| TRAIN_306 | NOVA Chemicals | 2022 | CHEMICALS | GRI_NEW | TRAIN |
| CTP_2023 | CTP | 2023 | CONSTRUCTION | ESRS | TEST_ESRS |
| HELVAR_2023 | HELVAR OY AB | 2023 | CONSTRUCTION | ESRS | TEST_ESRS |
| HH_2023 | H+H | 2023 | CONSTRUCTION | ESRS | TEST_ESRS |
| TRAIN_242 | Heidelberg Materials | 2022 | CONSTRUCTION | GRI_NEW | TRAIN |
| TRAIN_119 | NESTE | 2021 | ENERGY | GRI_OLD | TRAIN |
| TRAIN_218 | Fortis Inc. | 2022 | ENERGY | GRI_NEW | TRAIN |
| TRAIN_228 | FortisBC | 2022 | ENERGY | GRI_NEW | DEV |
| SANTADER_2023 | SANTADER BANK POLSKA GROUP | 2023 | FINANCIAL SERVICES | ESRS | TEST_ESRS |
| TRAIN_191 | YUANTA FINANCIAL HOLDINGS | 2021 | FINANCIAL SERVICES | GRI_OLD | TRAIN |
| TRAIN_194 | Banca Transilvania | 2020 | FINANCIAL SERVICES | GRI_OLD | TRAIN |
| TRAIN_239 | Gulf International Bank | 2022 | FINANCIAL SERVICES | GRI_NEW | DEV |
| TRAIN_307 | Taishin Financial Holding | 2021 | FINANCIAL SERVICES | GRI_OLD | TRAIN |
| TRAIN_71 | Capital One | 2021 | FINANCIAL SERVICES | GRI_NEW | TRAIN |
| TRAIN_127 | LOOMIS | 2022 | FINANCIAL SERVICES | GRI_NEW | TRAIN |
| TRAIN_155 | Loomis | 2021 | FINANCIAL SERVICES | GRI_OLD | TRAIN |
| TRAIN_0 | ALLY FINANCIAL | 2021 | FINANCIAL SERVICES | GRI_OLD | TRAIN |
| TRAIN_2 | Energy Recovery | 2021 | TECHNOLOGY | GRI_OLD | TRAIN |
| TRAIN_77 | Motorola Solutions | 2021 | TECHNOLOGY | GRI_NEW | TRAIN |
| TRAIN_3 | Meta | 2021 | TECHNOLOGY | GRI_OLD | TRAIN |
| KPN_2023 | KPN | 2023 | TELECOMMUNICATIONS | GRI_NEW | TEST_GRI |
| TRAIN_153 | NTT DOCOMO | 2020 | TELECOMMUNICATIONS | GRI_OLD | TRAIN |
| ARLA_2023 | ARLA | 2023 | CONSUMER PACKAGED GOODS | ESRS | TEST_ESRS |
| TRAIN_81 | Ryanair | 2022 | AVIATION | GRI_NEW | TRAIN |
| TRAIN_124 | HITEJINRO | 2023 | CONSUMER PACKAGED GOODS | GRI_NEW | DEV |
| TRAIN_212 | Molson Coors Beverage Company | 2022 | CONSUMER PACKAGED GOODS | GRI_OLD | TRAIN |
| TRAIN_197 | Illumina | 2021 | BIOTECH | GRI_OLD | TRAIN |
| TRAIN_181 | CWT | 2022 | LOGISTICS | GRI_OLD | TRAIN |
| KERRY GROUP_2023 | KERRY GROUP | 2023 | CONSUMER PACKAGED GOODS | GRI_NEW | TEST_GRI |
| LACTALIS_2023 | LACTALIS | 2023 | CONSUMER PACKAGED GOODS | GRI_NEW | TEST_GRI |
| TRAIN_138 | LS ELECTRIC | 2023 | ELECTRONICS | GRI_NEW | TRAIN |
| TRAIN_245 | TAIFLEX | 2023 | ELECTRONICS | GRI_NEW | TRAIN |
| TRAIN_185 | KONE | 2022 | MANUFACTURING | GRI_NEW | TRAIN |
| TRELLEBORG_2019 | Trelleborg AB | 2019 | MANUFACTURING | GRI_OLD | TRAIN |
| TRELLEBORG_2020 | Trelleborg AB | 2020 | MANUFACTURING | GRI_OLD | TRAIN |
| TRELLEBORG_2021 | Trelleborg AB | 2021 | MANUFACTURING | GRI_OLD | TRAIN |
| TRELLEBORG_2022 | Trelleborg AB | 2022 | MANUFACTURING | GRI_NEW | DEV |
| TRELLEBORG_2023 | Trelleborg AB | 2023 | MANUFACTURING | GRI_NEW | TEST_GRI |
| VANDEMOORTELE_2023 | Vandemoortele Group | 2023 | CONSUMER PACKAGED GOODS | ESRS | TEST_ESRS |
| AB SKF_2023 | SKF GROUP | 2023 | MANUFACTURING | GRI_NEW | TEST_GRI |
| TRAIN_137 | UNION STEEL HOLDINGS LIMITED | 2021 | MANUFACTURING | GRI_OLD | TRAIN |
| TRAIN_169 | If P&C Insurance | 2020 | FINANCIAL SERVICES | GRI_OLD | TRAIN |
| TRAIN_65 | Generali Group | 2022 | FINANCIAL SERVICES | GRI_OLD | TRAIN |
| TRAIN_116 | SK Inc. | 2022 | FINANCIAL SERVICES | GRI_OLD | TRAIN |
| TRAIN_90 | SK Inc. | 2023 | FINANCIAL SERVICES | GRI_NEW | TEST_GRI |
| TRAIN_223 | Investor AB | 2022 | FINANCIAL SERVICES | GRI_NEW | TRAIN |
| TRAIN_302 | EQT | 2022 | FINANCIAL SERVICES | GRI_NEW | TRAIN |
| SGL_2023 | SCAN GLOBAL LOGISTICS | 2023 | LOGISTICS | ESRS | TEST_ESRS |
| TRAIN_187 | Ferrexpo | 2020 | MINING | GRI_OLD | TRAIN |
| TRAIN_24 | Coeur Mining | 2022 | MINING | GRI_NEW | TRAIN |
| TRAIN_55 | The Metals Company | 2021 | MINING | GRI_NEW | TRAIN |
| TRAIN_9 | Methanex | 2021 | CHEMICALS | GRI_OLD | TRAIN |
| TRAIN_1 | KUMBRA IRON ORE LIMITED | 2021 | MINING | GRI_OLD | TRAIN |
| TRAIN_143 | KUMBRA IRON ORE LIMITED | 2020 | MINING | GRI_OLD | TRAIN |
| TRAIN_4 | Billerud | 2022 | MANUFACTURING | GRI_NEW | TRAIN |
| TRAIN_126 | ABBOTT | 2022 | PHARMA | GRI_NEW | TRAIN |
| TRAIN_20 | Pfizer | 2021 | PHARMA | GRI_OLD | TRAIN |
| TRAIN_13 | VASAKRONAN | 2020 | FINANCIAL SERVICES | GRI_OLD | TRAIN |
| TRAIN_66 | Dream Unlimited Corp. | 2021 | FINANCIAL SERVICES | GRI_NEW | TRAIN |
| TRAIN_225 | Green Plains | 2021 | ENERGY | GRI_OLD | TRAIN |
| TRAIN_70 | TJX Companies | 2022 | RETAIL | GRI_OLD | TRAIN |
| TRAIN_171 | MACRONIX INTERNATIONAL | 2021 | ELECTRONICS | GRI_OLD | TRAIN |
| TRAIN_170 | COUPA | 2022 | LOGISTICS | GRI_OLD | TRAIN |
| TRAIN_8 | Amer Sports | 2022 | RETAIL | GRI_NEW | TRAIN |
| TRAIN_75 | Everest Textile Co., Ltd. | 2021 | MANUFACTURING | GRI_OLD | TRAIN |

Table 5: Company names and years of the ESG reports in `ESG-CID`.

## E Dataset Example

In this section, we provide examples of the GRI index and the ESRS index from the HYUNDAI

2024 sustainability report. This communicates the complexity of the existing pdf data and why generating an ESRS report from the the GRI format report is challenging. Additionally, once relevent ESRS index and GRI index are identified; collating related content is non-trivial. See Figures 6, 7, and 8 for example content indices both in ESRS and GRI standards.

# ESRS (European Sustainability Reporting Standards)

## ESRS 2. General Disclosures

| Indicator No. | Title | Page |
|---|---|---|
| ESRS 2 BP-1 | General basis for preparation of the sustainability statements | 124 |
| ESRS 2 BP-2 | Disclosures in relation to specific circumstances | 28, 36, 42, 43, 97, 98, 100, 117-122 |
| ESRS 2 GOV-1 | The role of the administrative, management and supervisory bodies | 9, 21, 81-85 |
| ESRS 2 GOV-2 | Information provided to and sustainability matters addressed by the undertaking's administrative, management and supervisory bodies | 82, 85 |
| ESRS 2 GOV-3 | Integration of sustainability-related performance in incentive schemes | 9, 17, 20, 37, 59 |
| ESRS 2 GOV-4 | Statement on sustainability due diligence | 50-53, 67-69 |
| ESRS 2 GOV-5 | Risk management and internal controls over sustainability reporting[1] | - |
| ESRS 2 SBM-1 | Market position, strategy, business model(s) and value chain | 6-7, 25-26 |
| ESRS 2 SBM-2 | Interests and views of stakeholders | 11-13 |
| ESRS 2 SBM-3 | Material impacts, risks and opportunities and their interaction with strategy and business model(s) | 15-17 |
| ESRS 2 IRO-1 | Description of the processes to identify and assess material impacts, risks and opportunities | 14 |
| ESRS 2 IRO-2 | Disclosure Requirements in ESRS covered by the undertaking's sustainability statements | 110-112 |

Figure 6: ESRS 2. General Disclosures Content Index of Hyundai found on page 110 of their 2024 sustainability report. The Indicator No. represents the standard's identifier, Title is used as the query text for our RAG system, and Page gives us the gold standard location of the relevant pages for the query within the report.

## ESRS E1. Climate Change

| Indicator No. | Title | Page |
|---|---|---|
| ESRS E1-1 | Transition plan for climate change mitigation | 32 |
| ESRS E1-2 | Policies related to climate change mitigation and adaptation | 23-32 |
| ESRS E1-3 | Actions and resources in relation to climate change policies | 32, 37 |
| ESRS E1-4 | Targets related to climate change mitigation and adaptation | 24-26, 30-32, 38 |
| ESRS E1-5 | Energy consumption and mix | 98 |
| ESRS E1-6 | Gross Scopes 1, 2, 3 and Total GHG emissions | 36, 98 |
| ESRS E1-7 | GHG removals and GHG mitigation projects financed through carbon credits | 16, 31 |
| | Avoided emissions of products and services | 15, 27 |
| ESRS E1-8 | Internal carbon pricing[2] | - |
| ESRS E1-9 | Potential financial effects from material physical and transition risks and potential climate-related opportunities | 22, 33-35 |

Figure 7: ESRS E1. Climate Change: Content index of the climate change related topics found on page 110 of the Hyundai 2024 sustainability report. The Indicator No. represents the standard's identifier, Title is used as the query text for our RAG system, and Page gives us the gold standard location of the relevant pages for the query within the report.

# GRI Index

## Topic Specific Standards - Environmental

| GRI Standards | | Page |
|---|---|---|
| No. | Title | |
| 301-1 | Materials used by weight or volume | 42, 98 |
| 301-2 | Recycled input materials used | 42, 98 |
| 301-3 | Reclaimed products and their packaging materials | 42 |
| 302-1 | Energy consumption within the organization | 98 |
| 302-2 | Energy consumption outside of the organization | 36 |
| 302-3 | Energy Intensity | 98 |
| 302-4 | Reduction of energy consumption | 23-24 |
| 303-1 | Interactions with water as a shared resource | 42-43, 99 |
| 303-2 | Management of impacts related to wastewater | 43, 100 |
| 303-3 | Water withdrawal | 99 |
| 303-4 | Water discharge | 99 |
| 303-5 | Water consumption | 20, 42, 99 |
| 304-1 | Operational sites owned, leased, managed in, or adjacent to, protected areas and areas of high biodiversity value outside protected areas | 46-48 |
| 304-2 | Significant impacts of activities, products and services on biodiversity | 46-48 |
| 304-3 | Habitats protected or restored | 46-48 |
| 304-4 | IUCN Red List species and national conservation list species with habitats in areas affected by operations | 48 |

| GRI Standards | | Page |
|---|---|---|
| No. | Title | |
| 305-1 | Direct (Scope 1) GHG emissions | 36, 98 |
| 305-2 | Energy indirect (Scope 2) GHG emissions | 36, 98 |
| 305-3 | Other indirect (Scope 3) GHG emissions | 36, 98 |
| 305-4 | GHG emissions intensity | 36, 98 |
| 305-5 | Reduction of GHG emissions | 23-32 |
| 305-7 | Nitrogen oxides (NOx), sulfur oxides (SOx), and other significant air emissions | 100 |
| 306-1 | Waste generation and significant waste-related impacts | 40-43 |
| 306-2 | Management of significant waste-related impacts | 40-43 |
| 306-3 | Waste generated | 100 |
| 306-4 | Waste diverted from disposal | 43, 100 |
| 306-5 | Waste directed to disposal | 100 |
| 308-1 | New suppliers that were screened using environmental criteria | 67-68 |
| 308-2 | Negative environmental impacts in the supply chain and actions taken | 69 |

Figure 8: GRI Content Index for GRI 300: Topic Specific Standards - Environmental.